\documentclass{article}

\usepackage[preprint]{corl_2024} 
\usepackage{mymacros}
\usepackage{wrapfig}

\title{Future Success Prediction in \\Open-Vocabulary Object Manipulation Tasks \\Based on End-Effector Trajectories}

\author{
  Motonari~Kambara and Komei~Sugiura \\
  Keio University, Japan \\
  \texttt{\{motonari.k714, komei.sugiura\}@keio.jp} \\
}

\begin{document}
\maketitle


\vspace{-8mm}
\begin{abstract}
This study addresses a task designed to predict the future success or failure of open-vocabulary object manipulation. 
In this task, the model is required to make predictions based on natural language instructions, egocentric view images before manipulation, and the given end-effector trajectories. Conventional methods typically perform success prediction only after the manipulation is executed, limiting their efficiency in executing the entire task sequence. We propose a novel approach that enables the prediction of success or failure by aligning the given trajectories and images with natural language instructions. 
We introduce Trajectory Encoder to apply learnable weighting to the input trajectories, allowing the model to consider temporal dynamics and interactions between objects and the end effector, improving the model's ability to predict manipulation outcomes accurately. 
We constructed a dataset based on the RT-1 dataset, a large-scale benchmark for open-vocabulary object manipulation tasks, to evaluate our method. 
The experimental results show that our method achieved a higher prediction accuracy than baseline approaches.
\end{abstract}

\keywords{Task Success Prediction and Open-Vocabulary Object Manipulation}
\vspace{8pt}
\begin{figure}[h]
    \centering
    \includegraphics[width=\linewidth]{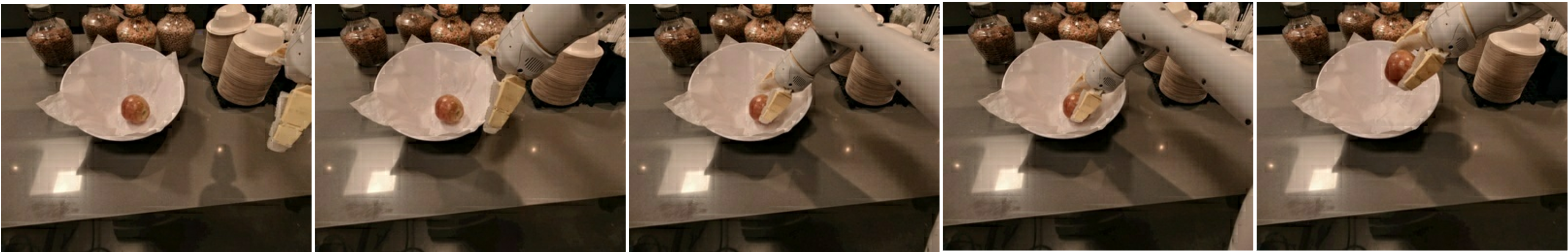}
    \vspace{-10pt}
    \caption{
Task Example. ``Pick an apple from the white bowl.'' is provided as a natural language instruction. In this example, the model is expected to predict 'Success' because the object manipulation is performed appropriately.
    }
\label{fig:task}
\end{figure}
\vspace{8pt}

\section{Introduction}

Object manipulation is essential in fields, such as household tasks~\citep{wu2023tidybot} and agriculture~\citep{lehnert2017autonomous, jun2021towards}.
Gaining insight into the success of this manipulation can improve its efficiency and safety because it avoids potential dangers caused by task failures and the unnecessary execution of tasks following object manipulation failures.

This study focuses on predicting the success of open-vocabulary object manipulations. The task involves predicting the success of the manipulations based on a given end effector's trajectory, an egocentric view of the image before the manipulation, and a natural language instruction sentence.
These task functions are challenging because they require considering future interactions between objects and the end effector based on the generated trajectory and their alignment with the natural language instructions.

A typical example of the task is depicted in Fig.~\ref{fig:task}. In this example, the instruction ``Pick an apple from the white bowl.'' is given. Given that the manipulator successfully grasps the apple from the white bowl, the model should predict a successful object manipulation.

Most existing methods related to manipulation success prediction perform the success/failure prediction after the manipulation execution. In contrast, predicting the success or failure of a task based on a given trajectory and an image can enable even more efficient execution of the entire task sequence.

In this study, we propose an open-vocabulary object manipulation success-prediction model. Our model predicts whether the object manipulation specified by an instruction sentence succeeds by aligning given trajectories and images before the manipulation. This alignment enables consideration of interactions between objects and the end effector from images and trajectories. Consequently, unlike many existing methods, our approach enables the prediction of the success or failure of object manipulation before the manipulation occurs. 

The contributions of this research are as follows:
\begin{itemize}
\item We propose a model that predicts whether the task specified by an instruction sentence can be appropriately executed by aligning the trajectory with the pre-manipulation image.
\item We introduce Trajectory Encoder to apply weighting to the trajectory using learnable parameters.
\end{itemize}

\section{Related Work}

Collision prediction during object manipulation is highly relevant to our work. For instance, post-collision decision strategies have been proposed (e.g., \citep{haddadin2017robot}). Moreover, several methods predict collisions using images and placement policies~\citep{mottaghi2016what, magassouba2021predicting, kambara2022relational}. Our method differs from these approaches by considering factors beyond collisions contributing to task failure.

Liu et al. proposed a task failure prediction model~\citep{liu2024model}. They proposed Failure Classifier to detect erroneous trajectory generation based on images and predicted trajectories. However, this model cannot process natural language instructions, making it unsuitable for direct application to the task addressed in this paper.

The proposed method is also closely related to subtask planning methods in long-horizon tasks~\citep{ishikawa2022moment, brohan2023saycan, brohan2022rt}. Some approaches determine the success or failure of a task after executing a subtask and replan subsequent subtasks based on that assessment~\citep{shirasaka2024selfrecovery, Driess2023palme, brohan2023saycan}. These methods are similar to the proposed method because they evaluate task success. 
For example, REFLECT~\citep{liu2023robofail} verifies whether predefined states are achieved based on target states for each object class.
However, unlike these methods that only detect failures during object manipulation, our proposed method predicts the success or failure of a manipulation task without relying on predefined target states. 
Moreover, our method enables more efficient execution by letting users know beforehand whether the manipulation will fail.

\section{Problem Statement}
This study focuses on Trajectory-based Manipulation Success Prediction (TMSP). In this task, a model should correctly predict whether an object manipulation will succeed or fail.
Fig.~\ref{fig:task} illustrates a typical example of the TMSP task. The leftmost image of Fig.~\ref{fig:task} presents the egocentric view image before object manipulation. The sequence of the images in Fig.~\ref{fig:task} indicates the trajectory of the end effector. The rightmost image illustrates the image after the manipulation. Suppose the instruction ``Pick an apple from the white bowl.'' is given. Because the object manipulation is performed successfully, the model is expected to predict `Success' based on the pre-manipulation image, the trajectory, and the instruction sentence.

In the TMSP task, the input consists of an egocentric image, the trajectory of the end effector, and a natural language instruction sentence.
The expected output is the predicted probability $p(\hat{y}=1)$, representing the likelihood that the manipulator will successfully perform the manipulation. $\hat{y}$ represents the predicted success or failure of the manipulation, with 1 indicating success.

\section{Proposed Method}

\begin{figure}[t]
    \centering
    \includegraphics[width=\linewidth]{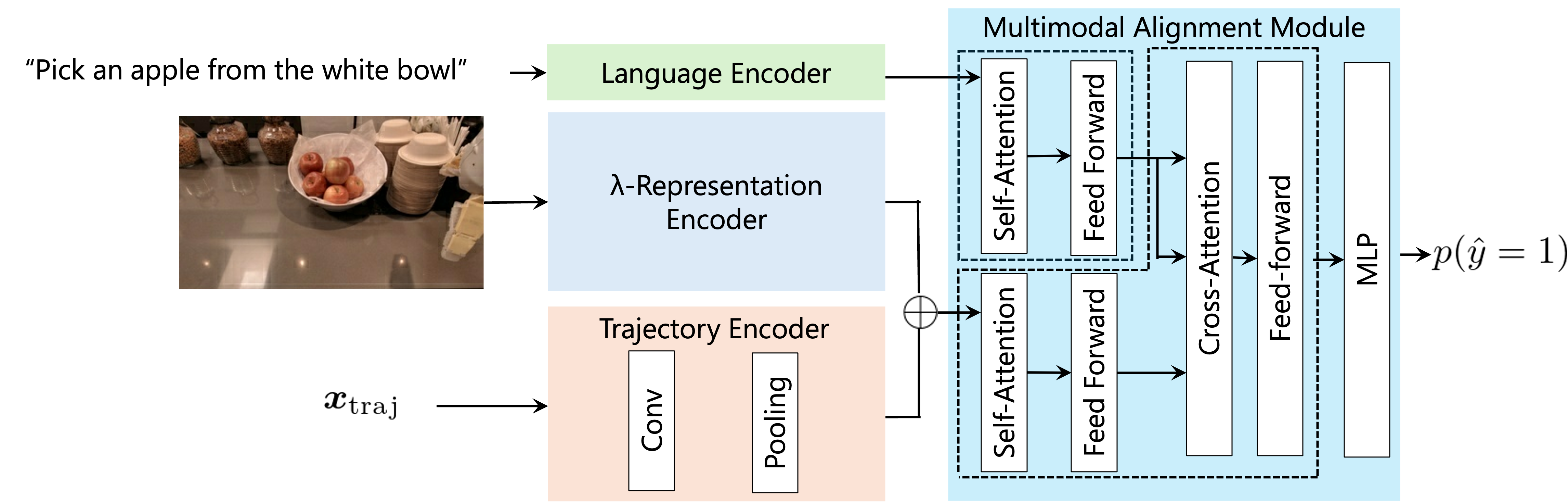}
    \vspace{-8pt}
    \caption{
Network overview of the proposed method. `Conv' and `MLP' represent the convolutional layer and multi-layer perceptron, respectively.
    }
    \label{fig:model}
\end{figure}

Fig. \ref{fig:model} shows an overview of the proposed method. The proposed method consists of Trajectory Encoder and $\lambda$-Representation Encoder. The input to the model is denoted as $\bm{x} = \{ \bm{x}_{\mathrm{txt}}, \bm{x}_{\mathrm{img}}, \bm{x}_{\mathrm{traj}} \}$, where $\bm{x}_{\mathrm{txt}}$, $\bm{x}_{\mathrm{img}}$, and $\bm{x}_{\mathrm{traj}}$ are the egocentric view image before object manipulation, the natural language instruction, and the end effector trajectory, respectively. $\bm{x}_{\mathrm traj}$ has a time series of length $T$. In the proposed method, the language feature $\bm{h}_{\mathrm{txt}}$ is first obtained from $\bm{x}_{\mathrm txt}$.

\subsection{Trajectory Encoder}
We introduce Trajectory Encoder, a module designed to filter the end effector's trajectory using learnable weights. This filtering process enables temporal trajectory compression for more efficient representation and processing.

The input to this module is $\bm{x}_{\mathrm{traj}} \in \mathbb{R}^{D \times T}$, and the output is $\bm{h}_{\mathrm{traj}}$, where $D$ denotes the number of degrees of freedom of the manipulator. This module is composed predominantly of convolutional and pooling layers. 

First, we obtain $\bm{x}_{\mathrm{conv}} \in \mathbb{R}^{D \times T}$ by applying a convolutional layer along the time dimension, capturing temporal patterns. The convolution focuses on modeling temporal dependencies in the trajectory data. 
We then apply a pooling layer along the time dimension of $\bm{x}_{\mathrm{conv}}$, reducing it from $T$ to $d_{\mathrm{trm}}$, resulting in a processed tensor $\bm{h}_{\mathrm{traj}} \in \mathbb{R}^{D \times d_{\mathrm{trm}}}$. This approach efficiently models the temporal dynamics of trajectories.

\subsection{$\lambda$-Representation Encoder}
Solving the TMSP task requires understanding the positional information of each object in the input image to predict whether the trajectory is interacting with the appropriate object. 

In the TMSP task, models are required to predict the success or failure of object manipulation based on open-vocabulary manipulation instructions, images, and trajectories. Thus, it is essential to ensure proper alignment with natural language to predict whether the manipulation instruction can be executed based on the images.

Therefore, we employ the $\lambda$-representation~\citep{goko2024task} as a visual feature aligned with natural language. This representation has been reported to be more effective than visual features obtained from the image encoder of CLIP, which is also aligned with natural language.

This module extracts the $\lambda$-representation $\bm{h}_{\lambda}$ from the input image. The input to this module is $\bm{x}_{\mathrm{img}}$, and the output is $\bm{h}_{\lambda}$. The process in this module is based on the $\lambda$-Representation Encoder module~\citep{goko2024task}. In contrast, for Narrative Representation, we use GPT-4o to generate a description about $\bm{x}_{\mathrm{img}}$.

We then compute cross-attention between $[\bm{h}_{\lambda} ; \bm{h}_{\mathrm{traj}}]$ and $\bm{h}_{\mathrm{txt}}$ using the $N$-layer transformer encoder-decoder. Finally, we obtain the predicted probability $p(\hat{y})$ of successful object manipulation.

\section{Experiments}

\subsection{Experimental Settings}
We constructed a dataset by extending the SP-RT-1 dataset~\citep{goko2024task}.
The SP-RT-1 dataset was built based on the RT-1 dataset~\citep{brohan2022rt}. The SP-RT-1 dataset includes images before and after object manipulation, object manipulation instruction sentences, and success/failure labels for each episode of object manipulation. 
In contrast, the TMSP task requires a dataset that includes egocentric view images before object manipulation, end effector trajectories, natural language instructions, and task outcome labels indicating success or failure.
Therefore, because the SP-RT-1 dataset generally contains the information required for the TMSP task, we constructed our dataset based on the SP-RT-1 dataset.

The end effector trajectories in each episode were lacking when using the SP-RT-1 dataset for the TMSP task. We adapted the trajectories included in the RT-1 dataset for the TMSP task by collecting the trajectories.
We split the dataset as follows in~\citep{goko2024task}.

The constructed dataset includes 13,915 samples, split into 11,915 training samples, 1,000 validation samples, and 1,000 test samples.
The manipulator used for data collection had a seven-dimensional action space~\citep{brohan2022rt}. 
An additional dimension was related to the opening and closing of the end effector.
Therefore, the trajectory at each timestep has eight dimensions.
 
We used an NVIDIA GeForce RTX 4090 with 24GB of VRAM, 64GB of RAM, and an Intel Core i9--13900KF. The training time for the proposed method was approximately 1.5 hours, and the inference time, excluding feature extraction, was 0.78 ms per sample.

\begin{wraptable}{r}{0.55\textwidth}
    \centering
    {
    \begin{tabular}{lc}
        \toprule
        Method & Accuracy [\%] \\
        \hline
        Contrastive $\lambda$-Repformer~\citep{goko2024task} & 74.9 $\pm$ 0.79 \\
        Ours w/o Traj. Enc. & $83.2$ $\pm$ 0.48 \\
        Ours & $\bm{83.4}$ $\pm$ 0.65 \\
        \hline
    \end{tabular}}
    \caption{
        Quantitative results. `Ours w/o Traj. Enc.' refers to the proposed method where Trajectory Encoder is replaced with a linear function. Bold indicates the accuracy with the highest value. 
    }
    \label{tab:quantitative}
\end{wraptable}

\subsection{Quantitative Results}
Table~\ref{tab:quantitative} presents the quantitative results, including the means and standard deviations. We conducted the experiments five times. In our experiments, we used Contrastive $\lambda$-Repformer as the baseline. This method was chosen because it demonstrated remarkable results in the SPOM task~\citep{goko2024task}, closely related to the TMSP task. Post-manipulation images were not provided to Contrastive $\lambda$-Repformer to prevent leakage of the object manipulation results.
Accuracy was used as the evaluation metric.

The accuracies of the baseline method and the proposed method were 74.9\% and 83.4\%, demonstrating that the proposed method outperformed the baseline by 8.5 points.
These results indicate that the proposed method considered the trajectory to predict the future success or failure of object manipulation.

\subsection{Qualitative Results}
\begin{figure}[t]
    \centering
    \includegraphics[width=\linewidth]{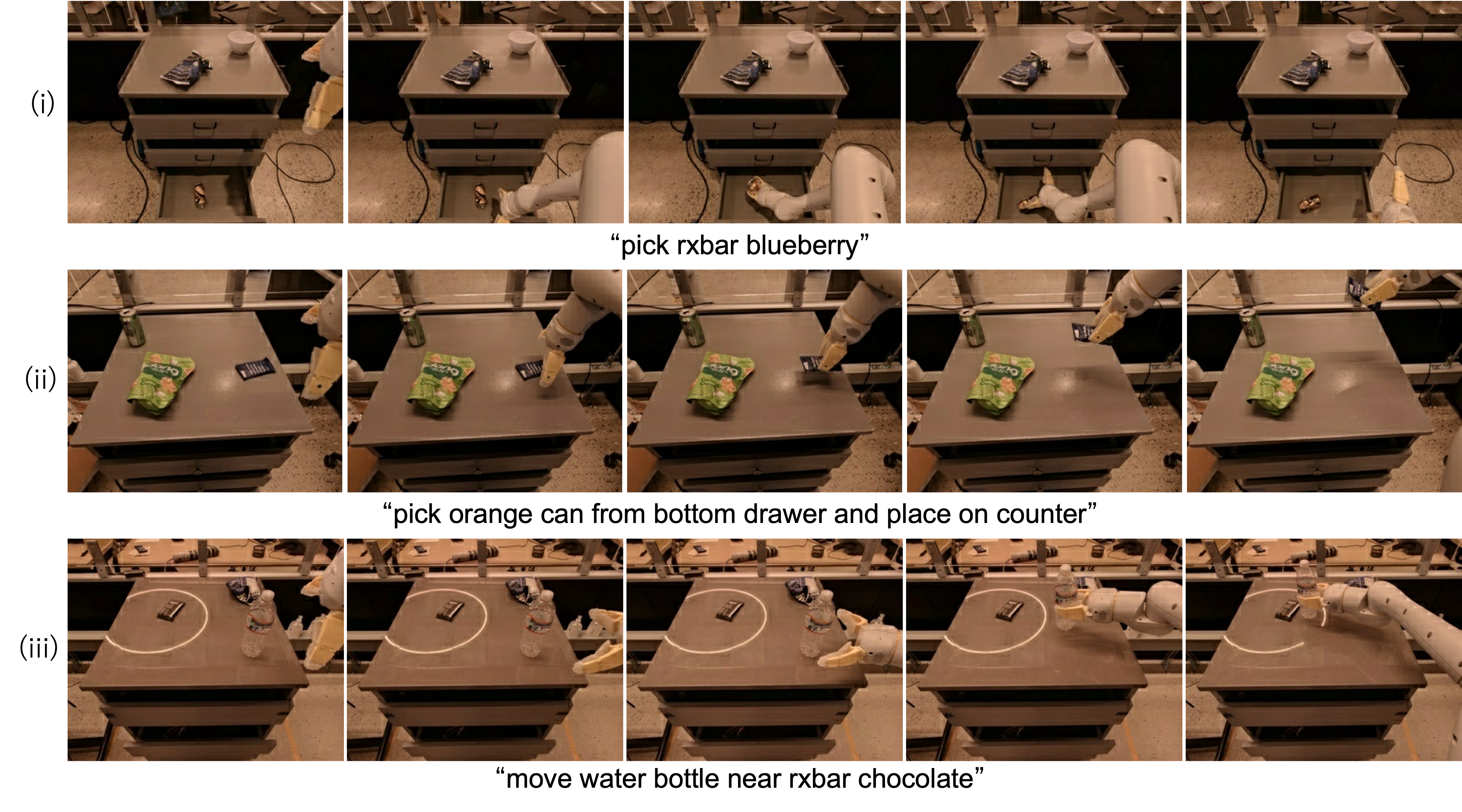}
    \vspace{-8pt}
    \caption{
    Qualitative results. Panels (i), (ii), and (iii) represent the True Positive, True Negative, and False Negative examples, respectively. The leftmost image in each panel illustrates the scene before manipulation.
    }
    \label{fig:qualitative}
\end{figure}
Fig.~\ref{fig:qualitative} illustrates the qualitative results. Panels (i), (ii), and (iii) are examples of True Positive, True Negative, and False Negative, respectively.

In Fig.~\ref{fig:qualitative}(i), the instruction was ``Pick rxbar blueberry.'' and the manipulator successfully grasped the protein bar. 
Therefore, the label for this episode was `Success'. The proposed method correctly predicted `Success', whereas the baseline method incorrectly predicted `Fail'.

In the episode depicted in Fig.~\ref{fig:qualitative}(ii), the instruction was ``Pick orange can from bottom drawer and place on counter.'' However, the manipulator failed to grasp the can and could not retrieve it from the drawer. Therefore, the label for this episode was `Fail'. The proposed method correctly predicted `Fail' for this episode. The baseline method incorrectly predicted `Success'. These results indicate that the proposed method could appropriately predict the success or failure of object manipulation by considering the end effector's trajectory and the positions of objects within the image.

In contrast, Fig.~\ref{fig:qualitative}(iii) illustrates an example where the proposed method made an incorrect prediction. In this episode, the instruction was ``Move water bottle near rxbar chocolate.'' 
The manipulator moved the chocolate bar close to the water bottle, so the label was `Success'. However, the proposed method predicted `Fail'. This episode was likely difficult because it necessitated accurately interpreting the spatial positions of multiple objects and aligning the trajectory with those positions.

\subsection{Ablation Study}
As an ablation study, we investigated the effectiveness of performance when Trajectory Encoder was removed. Table~\ref{tab:quantitative} also presents the results. In the table, `Ours w/o Traj. Enc.' represents the model where Trajectory Encoder was replaced with a linear function. The accuracies of Models (i) and (ii) were 83.2\% and 83.4\%, indicating a 0.2-point improvement using Trajectory Encoder. 
The simple-weighting approach applied within Trajectory Encoder produced a slight improvement.

\section{Conclusions}

In this study, we focused on a task to predict the future success or failure of open-vocabulary object manipulation, based on natural language instructions, egocentric view images before manipulation, and end effector trajectories.

Our contributions are as follows:
\begin{itemize}
\item We proposed a model that predicts whether the task specified by an instruction sentence can be appropriately executed by aligning the trajectory with the pre-manipulation image.
\item We introduced Trajectory Encoder, which applies weighting to the trajectory using learnable parameters.
\item The experimental results demonstrated that our method achieved higher prediction accuracy than the baseline method.
\end{itemize}

As noted in Section 5.4, introducing Trajectory Encoder is modestly beneficial.
Therefore, future research should consider improvements to Trajectory Encoder, such as using the trajectory as a visual prompt in the input images.


\clearpage
\acknowledgments{This work was partially supported by JSPS KAKENHI Grant Number 23K03478, JST Moonshot, and NEDO.}


\bibliography{main}  
\end{document}